\documentclass[runningheads]{llncs}
\usepackage[T1]{fontenc}

\usepackage{graphicx}
\usepackage{amssymb}
\usepackage{amsmath}
\usepackage{cleveref}
\usepackage{siunitx}
\usepackage{booktabs}
\usepackage{xfrac}
\usepackage{url}
\usepackage{subcaption}
\usepackage{graphicx}

\begin{document}
\title{NeuralPDE: Modelling Dynamical Systems from Data}
%
%

\author{Andrzej Dulny \and
Andreas Hotho \and
Anna Krause}
\authorrunning{A. Dulny et al.}
%
\institute{Chair for Computer Science X Data Science, University of Würzburg, Germany 
\email{\{andrzej.dulny,andreas.hotho,anna.krause\}@informatik.uni-wuerzburg.de}}

\maketitle              
\begin{abstract}
Many physical processes such as weather phenomena or fluid mechanics are governed by partial differential equations (PDEs).
Modelling such dynamical systems using Neural Networks is an active research field.
However, current methods are still very limited, as they do not exploit the knowledge about the dynamical nature of the system, require extensive prior knowledge about the governing equations or are limited to linear or first-order equations.
In this work we make the observation that the Method of Lines used to solve PDEs can be represented using convolutions which makes convolutional neural networks (CNNs) the natural choice to parametrize arbitrary PDE dynamics.
We combine this parametrization with differentiable ODE solvers to form the NeuralPDE Model, which explicitly takes into account the fact that the data is governed by differential equations.
We show in several experiments on toy and real-world data that our model consistently outperforms state-of-the-art models used to learn dynamical systems.

\end{abstract}
\section{Introduction}
Deep learning methods have brought revolutionary advances in computer vision, time series prediction and machine learning in recent years. 
Handcrafted feature selection has been replaced by modern end-to-end systems, allowing efficient and accurate modelling of a variety of data.
In particular, convolutional neural networks (CNNs)  automatically learn features on gridded data, such as images or geospatial information, which are invariant to spatial translation~\cite{Goodfellow2016}.
Recurrent neural networks (RNNs) such as long short-term memory networks (LSTMs) or gated recurrent units (GRUs) are specialised for modelling sequential data, such as time series or sentences (albeit now replaced by transformers)~\cite{Lipton2015}.

Recently, modelling dynamical systems from data has gained attention as a novel and challenging task~\cite{Karlbauer2019,Praditia2021,So2021,Li2020a}.
These systems describe a variety of physical processes such as weather phenomena~\cite{Rasp2020}, wave propagation~\cite{Karlbauer2019}, chemical reactions~\cite{Rudolph2005}, and computational fluid dynamics~\cite{BelbutePeres2020}.
All dynamical systems are governed by either ordinary differential equations (ODEs) involving time derivatives or partial differential equations (PDEs) involving time and spacial derivatives.
Due to their chaotic nature, learning such systems from data remains challenging for current models~\cite{Bronstein2017}. 
 
In recent years, several approaches to model dynamical data incorporating prior knowledge about the physical system have been proposed~\cite{Raissi2017,Praditia2021,Long2019,Berg2019}. 
However, most of the models make specific assumptions about the type or structure of the underlying differential equations:
they have been designed for specific problem types such as advection-diffusion problems, require prior knowledge about the equation such as the general form or the exact equation, or are limited to linear equations. 
In current literature only a handful of flexible approaches exist~\cite{Karlbauer2019,Ayed2019,Iakovlev2021}.

\begin{figure*}[t]
    \centering
    \includegraphics[width=1\textwidth]{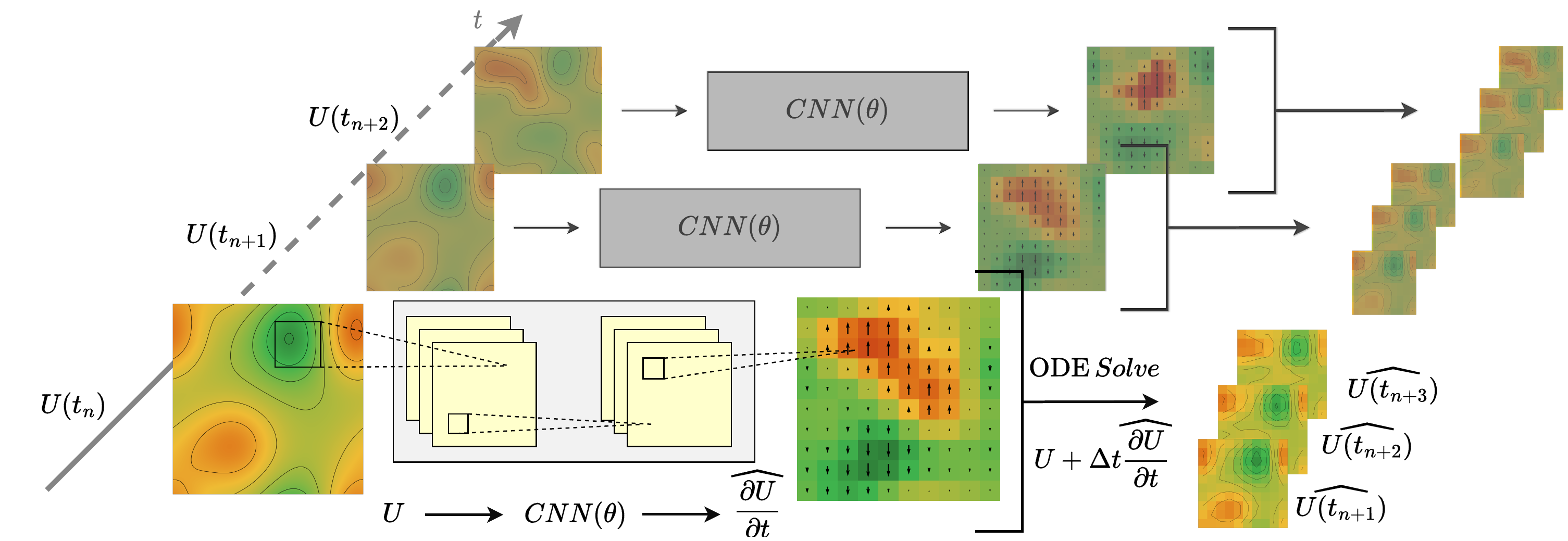}
    \caption{NeuralPDE: combining the Method of Lines and NeuralODE. Our model employs a CNN to parametrize the dynamics of the system $\frac{\partial\mathcal{U}}{\partial t}$. This allows the representation of the PDE by a system of ODEs (Method of Lines) which is solved using any differentiable ODE Solver predicting multiple future states (three in the figure above). The CNN is trained using adjoint backpropagation.}
    \label{fig:neuralPDE_overview}
\end{figure*}

In this work we propose NeuralPDE, a novel approach for modelling spatio-temporal data.
NeuralPDE learns the dynamics of partial differential equations using convolutional neural networks as summarized in~\Cref{fig:neuralPDE_overview}.
The derivative of the system is used to solve the underlying equations using the Method of Lines~\cite{Schiesser2012} in combination with differentiable ODE solvers~\cite{Chen2018}. 
Our approach works on an end-to-end basis, without assuming any prior constraints on the underlying equations, while taking advantage of the dynamical nature of the data by explicitly solving the governing differential equations.

The main contributions of our work are\footnote{Our code will be made publicly available upon publication.}:
\begin{enumerate}
    \item We combine NeuralODEs and the Method of Lines through usage of CNNs to account for the spatial component in PDEs.
    \item We propose using general CNNs that do not require prior knowledge about the underlying equations.
    \item NeuralPDEs can inherently learn continuous dynamics which can be used with arbitrary time discretizations.
    \item We demonstrate that our model is applicable to a wide range of dynamical systems, including non-linear and higher-order equations.
    \
\end{enumerate}

\section{Related Work}
\label{sec:related_work}

NeuralODEs~\cite{Chen2018} introduces continuous depth neural networks for parametrizing an ODE. 
The networks are combined with a standard ODE solver for solving the ODE.
NeuralODE forms the basis for our method in the same way that numerical ODE solvers are the basis for one family of numerical PDE solvers.

Many approaches for learning dynamical systems from data operate under strong assumptions about the underlying data:
Universal Differential Equations (UDE)~\cite{Rackauckas2020}, Physics Informed Neural Networks (PINN)~\cite{Raissi2017a}, and PDE Net 2.0~\cite{Long2019} require prior knowledge about the generating equations.
UDEs use separate neural networks to model each component of a PDE and have to be redesigned manually for every new PDE.
PINNs are a machine learning technique for neural networks which design the loss function such, that it satisfies the initial value problem of the PDE.
PDE-Net 2.0 assumes a library of available components and learns the parameters of the linear combination of these components using a ResNet-like model.
Finite Volume Networks (FINN)~\cite{Praditia2021} integrate the finite volume method with neural networks, but are strictly limited to advection-diffusion type equations.
Our results show that none of the restrictions apply to NeuralPDE: we do not need to know the exact PDE that governed the data and make no assumption about the structure of the governing PDE.

Flexible approaches include Distana~\cite{Karlbauer2019}, hidden state models~\cite{Ayed2019}, and the approaches proposed by Berg~\cite{Berg2019} and Iakovlev~\cite{Iakovlev2021}. 
Distana~\cite{Karlbauer2019} describes a neural network architecture that combines two types of LSTM-based kernels: 
predictive kernels make predictions at given spatial positions, transitional kernels model transitions between adjacent predictive kernels.
Distana proved successful in modeling wave equations and is applicable for further problems.
Iakovlev et al.~\cite{Iakovlev2021} propose using message passing graph neural networks in conjunction with Neural ODEs for modelling non-equidistant spatial grids and non-constant time intervals and evaluate their method on generated data.
In contrast to our approach, they use message passing graph neural networks which are inherently computationally less efficient than our method.
We provide a theoretical justification for using convolutional filters and use real-world as well as generated data for our experiments.

Berg~\cite{Berg2019} introduce a two step procedure: in the first step, the data is approximated by an arbitrary model.
In the second step a differentiation operator is approximated by training a neural network on the data approximator and its derivatives up to a given order.

Ayed et al.~\cite{Ayed2019} introduce the hidden state method with a learnable projection matrix to transform observed variables into a hidden state.
The authors apply their method to training small ResNets as parametrizations of dynamics on toy data as well as real world data sets.
Contrary to their method, we do not assume an underlying hidden process and instead directly learn the dynamic. 
Additionally we do not use residual connections in our parametrization, as our theoretical results show (\Cref{sec:our_method}) that direct convolutions are the best choice.

\section{Task}
%
Dynamical systems can be defined as a deterministic rule of evolution of a state in time~\cite{Kuznetsov1995}.
At any point in time $t\in T$ the entirety of the system is assumed to be completely described by a set of space variables $x$ from the state space $X$. 
The evolution of the system is given by the evolution function: 
\begin{equation}
    \Phi\colon T\times X \longrightarrow X
\end{equation}
which describes the how an initial state $x_0\in X$ is transformed into the state $x_1\in X$ after time $t_1\in T$ as $\Phi(t_1, x_0) = x_1$.
An important property of dynamical systems is their time homogeneity, meaning the evolution of the state only depends on the current state:
\begin{equation}
    \Phi(t_1, \Phi(t_2, x)) = \Phi(t_1+t_2, x)
\end{equation}

The main concern of this work is dynamical systems governed by a set of partial differential equations.
These are continuous spatio-temporal systems where the state at each point in time is described by a field of $k$ quantities on a given spatial domain $\Omega \subseteq \mathbb{R}^n$.
Examples of dynamical systems that can be described by PDE include many physical systems such as weather phenomena~\cite{Rasp2020} or wave propagation~\cite{Karlbauer2019}.
These systems often exhibit chaotic behaviour which makes them difficult to model with classical machine learning models~\cite{Iooss1983}.

We define the task of \textit{modeling dynamical systems from data} as a spatio-temporal time series prediction task, where from one or more states used as input the model should predict the evolution of the state for the next $H$ timesteps.
As opposed to physical simulations (usually used to model such systems) where the governing equation is known, in this task the equation is assumed to be unknown.
Additionally, retrieving the exact form of the equation is also not part of the task, which is the task of \textit{learning differential operators from data}~\cite{Long2019}.

\section{Neural PDE}
\label{sec:our_method}
In this section we describe our method, which combines NeuralODEs and the Method of Lines through the use of a multi-layer convolutional neural network to model arbitrarily complex PDEs.
Our primary focus lies on modelling spatio-temporal data describing a dynamical system and not on recovering the exact parameters of the differential equation(s). 

\subsection{Method of Lines}
The Method of Lines describes a numerical method of solving PDEs, where all of the spatial dimensions are discretized and the PDE is represented as a system of ordinary differential equations of one variable, for which common ODE solvers can be applied~\cite{Schiesser2012}. 
Given a partial differential equation of the form
\begin{equation}\label{eq:pde_system}
    \frac{\partial u}{\partial t} = f(t, u, \frac{\partial u}{\partial x}, \frac{\partial u}{\partial y}, \ldots)
\end{equation}
where $u = u(t, x, y), x\in X, y\in Y$ is the unknown function, the spatial domain $X\times Y$ is discretized on a regular grid $X\sim\{x_1, x_2, \ldots, x_N\}$ and $Y\sim\{y_1, y_2, \ldots, y_M\}$.
The function $u$ can then be represented as $N\cdot M$ functions of one variable (i. e. time):
\begin{equation}\label{eq:pde_discretization}
    u(t) \simeq 
    \left[ \begin{array}{rrr}
        u(t, x_1, y_1) & \cdots & u(t, x_N, y_1) \\ 
        \vdots & \ddots & \vdots \\
        u(t, x_1, y_M) & \cdots & u(t, x_N, y_M) \\ 
    \end{array}\right]
     =\colon \mathcal{U}
\end{equation}
From this representation one can derive the discretization of the spatial derivatives:
\begin{equation}
    \frac{\partial u}{\partial x}(t, x_i, y_i) = \frac{u(t, x_{i+1}, y_i)-u(t, x_{i-1}, y_i)}{x_{i+1}-x_{i-1}}
\end{equation}
and
\begin{equation}
    \frac{\partial u}{\partial y}(t, x_i, y_i) = \frac{u(t, x_i, y_{i+1})-u(t, x_i, y_{i-1})}{y_{i+1}-y_{i-1}}
\end{equation}
When a fixed grid size is used for the discretization, the spatial derivatives can thus be represented as a convolutional operation~\cite{Goodfellow2016}:
\begin{equation}\label{eq:partial_derivatives}
    \mathcal{U}_x
    = conv(\frac{1}{2\Delta x}
    \left[ \begin{array}{rrr}
        0 & 0 & 0 \\ 
        -1 & 0 & 1 \\
        0 & 0 & 0 \\ 
    \end{array}\right],
    \mathcal{U}) \qquad
    \mathcal{U}_y
    = conv(\frac{1}{2\Delta y}
    \left[ \begin{array}{rrr}
        0 & -1 & 0 \\ 
        0 & 0 & 0 \\
        0 & 1 & 0 \\ 
    \end{array}\right],
    \mathcal{U})
\end{equation}
Where $\Delta x$ and $\Delta y$ are the constant grid sizes for both spatial dimensions:
\begin{equation}
    \begin{aligned}
        \Delta x  &= x_{i+1}-x_{i}, i = 1, \ldots, N \\
        \Delta y  &= y_{i+1}-y_{i}, i = 1, \ldots, M
    \end{aligned}
\end{equation}
Higher-order spatial derivatives can be represented in a similar fashion by a convolutional operation on the lower-order derivatives.
This can be easily seen from the representation
\begin{equation}
    \frac{\partial^{p+q} u}{\partial x^p \partial y^q} = \frac{\partial}{\partial x} \frac{\partial^{p+q-1}u}{\partial x^{p-1}\partial y^q} = \frac{\partial}{\partial y} \frac{\partial^{p+q-1}u}{\partial x^{p}\partial y^{q-1}}
\end{equation}
as higher-order derivatives are defined as derivatives of lower-order derivatives.

The original PDE can now be represented as a system of ordinary differential equations, each representing the trajectory of a single point in the spatial domain (thus the name Method of Lines):

\begin{equation}\label{eq:ode_system}
    \frac{d\mathcal{U}}{dt} \simeq  f(t, \mathcal{U}, \mathcal{U}_x, \mathcal{U}_y, \ldots)=f^*(t, \mathcal{U})
\end{equation}
for which any numerical ODE solver can be used.

\subsection{NeuralPDEs}
Our method makes the assumption that the spatio-temporal data to be modelled is governed by a partial differential equation of the form \Cref{eq:pde_system}, but by physical constraints of the measuring process, the data has been sampled on a discrete spatial grid as in~\Cref{eq:pde_discretization} and depicted in~ \Cref{fig:neuralPDE_overview} on the bottom left.
We also assume that the dynamics of the system only depends on the state of the system itself
\begin{equation}
    f^*(t, \mathcal{U}) = f^*(\mathcal{U})
\end{equation}

As can be seen from \Cref{eq:partial_derivatives}, the spatial derivatives of the discretized PDE can be represented by a convolutional filter on the values of $\mathcal{U}$ and thus the whole dynamics of the system (which depends on the spatial derivatives) can be recovered from $\mathcal{U}$. 

\Cref{fig:neuralPDE_overview} shows an overview of our model.
Given the state of the system $\mathcal{U}_0$ at $t=t_0$, our method uses the Method of Lines representation of the underlying PDE (given by \Cref{eq:ode_system}) and employs a multi-layer convolutional network to parametrize the unknown function~$f^*$ describing the dynamics of the system
\begin{equation}
    \frac{d\mathcal{U}}{dt} \simeq  f^*(\mathcal{U}) \simeq \textrm{CNN}_\theta(\mathcal{U})
\end{equation}

Similar to NeuralODEs~\cite{Chen2018}, the parametrization of the dynamics is used in combination with differentiable ODE solvers.
Predictions are made by numerically solving the ODE Initial Value Problem given by
\begin{equation}
    \begin{split}
        &\frac{d\mathcal{U}}{dt} = \textrm{CNN}_\theta(\mathcal{U}) \\
        &\mathcal{U}(t_0) = \mathcal{U}_0
    \end{split}
\end{equation}
for time points $t_1, \ldots ,t_K$.
The weights $\theta$ of the parametrization $\textrm{CNN}_\theta$ are updated using adjoint backpropagation as described in~\cite{Chen2018}. 

For higher-order equations our model is augmented with additional channels corresponding to higher order derivatives.
Given the ordinary differential equation system
\begin{equation}
    \frac{d^p \mathcal{U}}{dt^p} = f^*(t, \mathcal{U})
\end{equation}
we parametrize the lower-order derivatives as separate variables

\begin{equation}
        \frac{d \mathcal{V}_{1}}{dt} := \frac{d \mathcal{U}}{dt} \qquad
        \frac{d \mathcal{V}_{2}}{dt} := \frac{d^2 \mathcal{U}}{dt^2} \qquad
        \cdots \qquad
        \frac{d \mathcal{V}_{p-1}}{dt} := \frac{d^{p-1} \mathcal{U}}{dt^{p-1}}
\end{equation}

Using these auxiliary variables $\mathcal{V}_{1}, \ldots \mathcal{V}_{p-1}$, the original equation \Cref{eq:higher_order_ode} can be rewritten as a system of $p$ first-order ODEs:

\begin{equation}
    \label{eq:higher_order_ode}
    \frac{d \mathcal{U}}{dt} = \mathcal{V}_{1} \qquad 
    \frac{d \mathcal{V}_{1}}{dt} = \mathcal{V}_{2} \qquad
    \cdots \qquad
    \frac{d \mathcal{V}_{p-1}}{dt} = f^*(t, \mathcal{U})
\end{equation}

We implement this augmentation method within NeuralPDE to represent higher-order dynamics.

\section{Data}
\label{sec:data}
Our aim for NeuralPDE is to be applicable to the largest possible variety of dynamical data.
For this, we curated a list of PDEs from related work as toy data, one simulated climate data set (PlaSim), and two reanalysis data sets (Weatherbench and Ocean Wave).

\paragraph{Toy Data Sets.}
We use several equation systems that are available from other publications as toy data sets: the advection-diffusion equation (AD), Burger's equation (B), the gas dynamics equation (GD), and the wave propagation equation (W).
The equation systems and the parameters used for data generation are available from appendix~A.
We use 50 simulations for different initial conditions for training, and 10 for validation and testing each.

\paragraph{Weatherbench~\cite{Rasp2020}.}
Weatherbench is a curated benchmark data set for learning medium-range weather forecasting model from data.
The data is derived from ERA5 archives and is accompanied by evaluation metrics, and several baseline models.
Instead of the very large raw data set, we use the data set with a spacial resolution of $5.625^{\circ}$ or $32\times 64$.
Following the recommendation of Rasp et al.~\cite{Rasp2020}, we use \textit{geopotential at 500 hPa pressure} and \textit{temperature at 850 hPa pressure} as target variables.
Data from years 1979 to 2014 is used for training, 2015 and 2016 for validation and 2017 and 2018 for testing.

\paragraph{Ocean Wave\protect\footnotemark.}
\refstepcounter{footnote}
\footnotetext{\url{https://resources.marine.copernicus.eu/product-detail/GLOBAL_MULTIYEAR_WAV_001_032/INFORMATION}}
\addtocounter{footnote}{-1}
The Ocean Wave data set contains aggregated global data on ocean sea surface waves from 1993 to 2020.
The data is on an equirectangular grid with a resolution of $1/5^{\circ}$ or approximately $20\,\mathrm{km}$ and with a temporal resolution of $3\,\mathrm{h}$.
We regrid the data to a spatial resolution of $32\times 64$ to match Plasim and Weatherbench.
We use \textit{spectral significant wave height (Hm0)}, \textit{mean wave from direction (VDMR)} and \textit{wave principal direction at spectral peak (VPED)} as target variables.
Data from years 1993 to 2016 is used for training, 2017 and 2018 for validation and 2019 and 2020 for testing.

\paragraph{PlaSim\protect\footnotemark.}
\refstepcounter{footnote}
\footnotetext{\url{https://www.mi.uni-hamburg.de/en/arbeitsgruppen/theoretische-meteorologie/modelle/plasim.html}}
\addtocounter{footnote}{-1}
The Planet Simulator (PlaSim) is a climate simulator using a medium complexity general circulation model for education and research into climate modelling and simulation.
For simulation, we used the setup \textit{plasimt21} as presented in~\cite{Scher2019}, Sec. 2.1.
Our simulation data contains one data point per day for 200 years.
We use temperature, geopotential, wind speed in x direction and wind speed in y direction at the lowest level of the simulation as our target variables.
Data from the first 180 years of the simulation is used for training, the 10 following years for validation and the years 191 through 200 for testing.

\section{Experiments}
We train and evaluate NeuralPDE and all selected comparison methods on the seven datasets as described below.

\begin{figure}[t]
    \centering
    \includegraphics[width=0.8\textwidth]{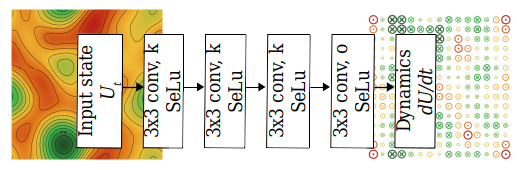}
    \caption{Architecture of our NeuralPDE models. We use four convolutional layers with $k=16$ channels, $3\times 3$ kernels, SeLu activation functions, and $o$ as the number of outputs.}
    \label{fig:neuralpde_architecture}
\end{figure}

\subsection{NeuralPDE Architecture}
\Cref{fig:neuralpde_architecture} shows the NeuralPDE architecture: a four layer CNN.
The first convolutional layer increases the number of channels to $k$, the last convolutional layer reduces the number of channels down to the number of inputs.
Then any number of intermediate layers each with $k$ channels can be used to perform the main computations.
After some primary experimentation we set the number of intermediate layers to 4 and the number of channels $k$ to 16.
The number of outputs $o$ depends on the choice of equation.
We train and evaluate two versions of our model using a first-order and second order dynamic as described in~\Cref{eq:higher_order_ode}. 
We denote these models as \textit{NeuralPDE-1} and \textit{NeuralPDE-2} respectively

\subsection{Comparison models}
We evaluate our model against several models from related work and simple baselines.
We follow \cite{Karlbauer2019} in the selection of our comparison models which we shortly describe in this section.
We omit models discussed in~\Cref{sec:related_work} which require prior knowledge about the equations.

\paragraph{Baseline.}
\textit{Persistence} refers to a model that directly returns it input as output.
It always takes the state at $t-1$ as the current prediction.

\paragraph{CNN.}
Similar to~\cite{Karlbauer2019} we use a CNN~\cite{LeCun1990} consisting of multiple convolutional layers as a comparison model.
We use the same architecture as for our NeuralPDE model.

\paragraph{ResNet.}
Motivated by the recent success of ResNet type architectures for modelling weather data~\cite{Rasp2021}, we include a simple ResNet model using identity mappings as proposed in~\cite{He2016}.
He et al.~\cite{He2016} use an residual unit consisting of BatchNorm layer followed by a ReLu activation, linear layer, BatchNorm, ReLu and another linear layer, which is then connected to the input by an additive skip-connection.
We stack 4 residual blocks preceded and followed by a linear CNN layer during our experiments.

\paragraph{Distana.}
\cite{Karlbauer2019} propose the distributed spatio-temporal artificial neural network architecture (DISTANA) to model spatio-temporal data.
Their model uses a graph network with learnable prediction kernels (LSTMs) at each node to learn spatio-temporal data.
We adopt the implementation of Distana from~\cite{Praditia2021}.

\paragraph{ConvLSTM.}
The convolutional LSTM as proposed in~\cite{Xingjian2015} replaces the fully connected layers within the standard LSTM model~\cite{Hochreiter1997} with convolutional layers.
It is well suited for modelling sequential grid data such as sequences of images~\cite{Wang2018}, or precipitation nowcasting~\cite{Xingjian2015}. 
We thus reason it might provide a strong comparison for modelling dynamical data.
We stack 4 ConvLSTM layers with 16 channels preceded and followed by a linear CNN layer for our experiments.

\paragraph{PDE-Net.} PDE-Net 2.0~\cite{Long2019} is a model explicitly designed to extract governing PDEs from data. 
Contrary to our approach it focuses on retrieving the equation in interpretable, closed form and not on modelling the data accurately. 
It uses a collection of learnable convolutional filters, connected together within a symbolic polynomial network to parametrize the dynamic. 
Our implementation is adapted from Long et al.~\cite{Long2019} and we use their parameters for our experiments.

\paragraph{Hidden State.} Ayed et al.~\cite{Ayed2019} propose a hidden state model, using a learnable projection to transform the input data into a higher-dimensional hidden state, where similarily to our approach, a differentiable solver is used to predict the next states. 
The predictions are projected again into the observed space by taking the first $o$ dimensions, where $o$ is the number of observed variables. 
We adopt the original parameters from~\cite{Ayed2019} to perform our experiments.
We project the observed data into a hidden state of 8 channels.

\subsection{Training}
Each model is trained in a closed-loop setting, where only the state of the system at $t_0$ is used as input for each of the models and the output $\hat{\mathcal{U}}_{t}$ at step $t$ is fed again into the model to make the prediction at step $t+1$.
For the higher-order models we initialize the higher-order derivatives as zeroes.

We train all our models using a horizon of $4$ time steps with batch size \num{8}, \num{5000} steps per epoch, and \num{5} epochs in total for both our \textit{NeuralPDE} and the \textit{Hidden State} model and \num{20} epochs for all other models.
We use the Adam optimizer with the learning rate of \num{0.001}.

All experiments are performed on a machine with a Nvidia RTX GPU, 16 CPUs and 32GB RAM.

\section{Results}
All models are evaluated using a prediction horizon of 16 time steps, using a hold-out test set as described in~\Cref{sec:data}. 
\Cref{tab:results_all} compares the RMSE averaged over 16 prediction steps and all target variables. Bold entries denote the best model for any given dataset.

\begin{table*}[t]
    \centering
    \label{tab:results_all}
    \setlength{\tabcolsep}{0.5em}
    \begin{tabular}{lccc}
    \toprule
     & \rotatebox[origin=c]{-90}{Ocean Wave}  
     & \rotatebox[origin=c]{-90}{Weatherbench}  
     & \rotatebox[origin=c]{-90}{PlaSim}  \\
    model       &                      &           &               \\
    \midrule

    persistence &                          0.558 &           0.114 &           0.708 \\
    CNN         &                          0.440 &           0.107 &           0.573 \\
    Distana     &                          0.440 &           0.108 &           0.559 \\
    ConvLSTM    &                          0.463 &           0.107 &           \underline{0.546} \\
    ResNet      &                 \textbf{0.427} &           0.103 &  \textbf{0.537} \\
    PDE-Net      &                 0.488 &           0.100 &           1.802 \\
    Hidden State &                          0.482 &  \textbf{0.096} &           0.572 \\
    NeuralPDE (Ours)  &                          \underline{0.435} &           \underline{0.097}  &           0.563 \\
    \bottomrule
    \end{tabular}
\end{table*}

The first four datasets (AD, B, GD, W) represent generated toy datasets of four different partial differential equations. 
Our model achieves state-of-the-art performance on all of these datasets except on the advection-diffusion equation, where the PDE-Net 2.0 model~\cite{Long2019} outperforms all other models by a large margin.
We hypothesize that the very simple dynamic governing this equation (given by just one linear convolutional filter) makes it very easy for the explicit approach used by the PDE-Net model to learn the dynamic.
On the other hand, our approach, which parametrizes the dynamic by a multilayer convolutional network is better at learning more complex systems of equations.

On the real-world datasets (OW, WB, P) NeuralPDE-1 closely matches the best state-of-the-art models on the Oceanwave and Weatherbench datasets coming in second best.
The Plasim dataset shows to be particularily difficult to learn for methods which directly parametrize the underlying dynamics (Hidden State, PDE-Net 2.0, NeuralPDE).
Our results show that the ResNet model achieves best performance on this dataset.
We hypothesize that the large time steps of 1 day in the simulated data makes it difficult for a continuous dynamic to be learned by our model.

\begin{figure*}[t]
    \centering
    \includegraphics[width=1\textwidth]{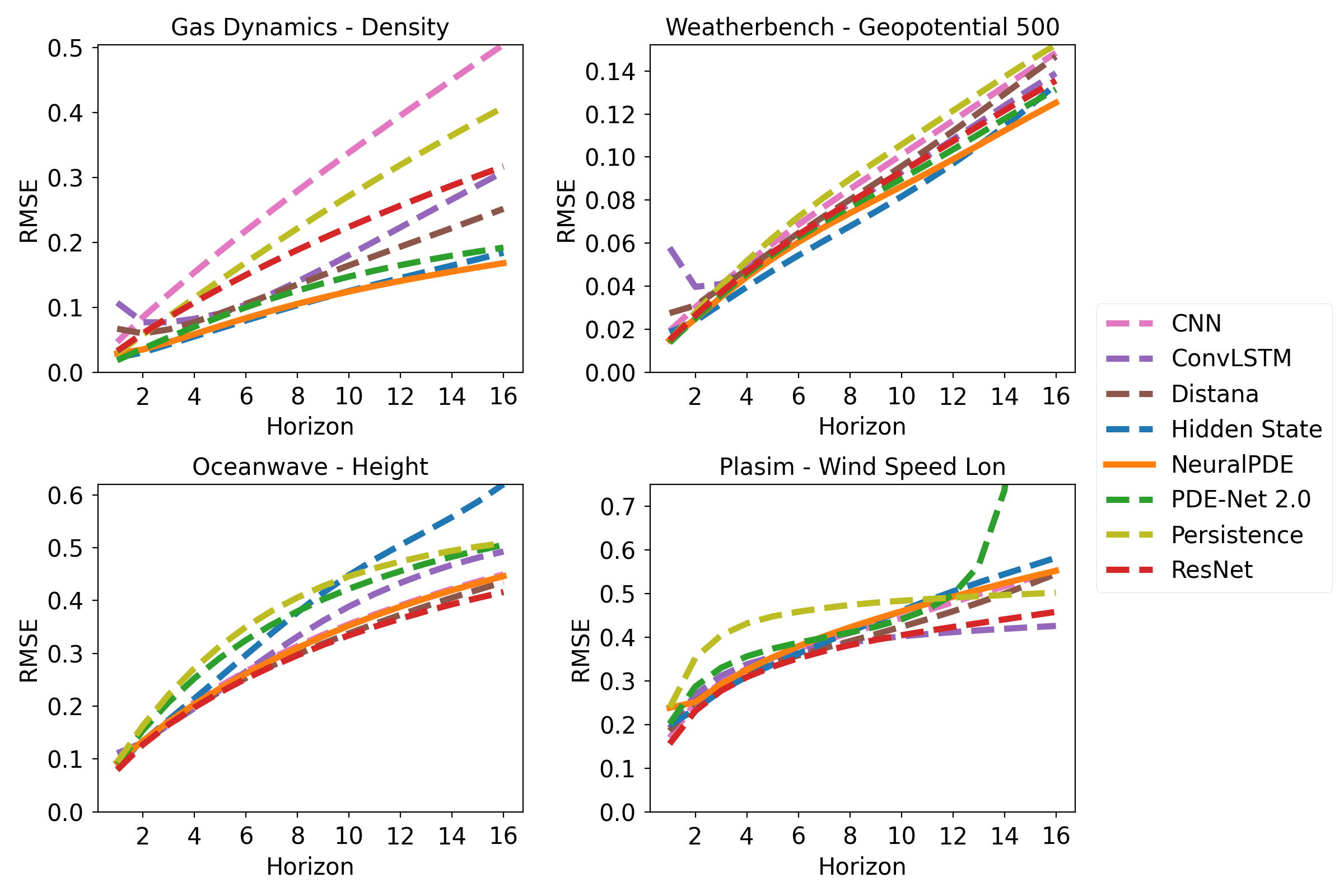}
    \caption{Predictions over different horizons. The figure shows the RMSE for four different datasets and target variables for all tested models as a function of the prediction horizon.}
    \label{fig:predictions_horizon}
\end{figure*}

\Cref{fig:predictions_horizon} shows the comparison of all tested models over increasing prediction horizons.
We only show a selection of different datasets and target variables, the full overview is available from Appendix~B. 
For all models the prediction accuracy decreases with increasing prediction horizon.

\section{Discussion}
Our method uses a multi-layer convolutional network as a generalized approach to represent differential equations.
Our experiments demonstrate that the same architecture can be applied successfully to learn a wide variety of PDE types, including linear and non-linear equations, equations in one and two dimensions, second-order equations, and coupled PDE systems of up to four equations.
In our current setting, NeuralPDE achieves state-of-the-art perfomance on generated data except for very simple equations, where we hypothesize a much simpler and less overparametrized network might perform better.
On the real-world datasets models that do not approximate the dynamic directly (ResNet) outperform our model and other models of this type, albeit not by a large margin.

One advantage of NeuralPDE over other flexible approaches is its inherent ability to directly capture the continous dynamics of the system.
While Distana or ResNet~\cite{Rasp2021} can only make discrete predictions at the next point in time, NeuralPDE can make predictions for any future point in time.
This also enables the modelling of data sampled at non-equidistant points in time.
In our experiment we used a fixed-step Euler solver, but in principle our method can be applied with any black-box numerical solver, including adaptive solvers like the Dormant-Prince (dopri) family of solvers~\cite{DORMAND198019}.





Currently, NeuralPDEs only encompass periodic boundary conditions.
We hypothesize that NeuralPDEs can be extended to other boundary conditions by adapting the parameterization of the convolutional layer, e.g. different padding types.
Moreover, the boundary conditions need to be specified beforehand and cannot yet be learned directly from data.

The Method of Lines comes with its own set of limitations: most prominently, it cannot be used to to solve elliptical second-order PDEs.
These limitations apply directly to NeuralPDEs as well.

Our model is a black box model that comes with limited interpretability.
While we do not directly learn the parametrization of a PDE, we could in theory extract the trained filters from the network for simple linear equations similarily to the PDE-Net~\cite{Long2019}.
However, as the system of equations grows more complex, the exact form of the PDE cannot be recovered from the learned weights.

If the order of the underlying system of equations is known, the appropriate order of our model can be chosen.
This is unfortunately not the case for many real-world applications.
However, as our experiments show, the first order model is a good first choice for a wide range of datasets.

\section{Conclusion}
In this work we proposed a novel approach to modelling dynamical data.
It is based on the Method of Lines used as a numerical heuristic for solving Partial Differential Equations, by approximating the spatial derivatives using convolutional filters.
In contrast to other methods, NeuralPDE does not make any assumptions about the structure of the underlying equations. Instead they rely on a deep convolutional neural network to parametrize the dynamics of the system.
We evaluated our method on a wide selection of dynamical systems, including non-linear and higher-order equations and showed that it is competetive compared to other approaches.

In our future work, we will address the remaining limitations:
First, we are planning to adapt NeuralPDE to learn boundary conditions from data.
Second, we are going to investigate combining other methods to model spatial dynamics with neural networks.
This includes other arbitrary mesh discretization methods as well as methods for continuous convolutions which could replace discretization completely.

\newpage
\bibliographystyle{splncs04}
\bibliography{references}

\end{document}


\title{NeuralPDE: Modelling Dynamical Systems from Data - Appendix}


\author{Andrzej Dulny \and
Andreas Hotho \and
Anna Krause}

\authorrunning{A. Dulny et al.}

\institute{Chair for Computer Science X Data Science, University of Würzburg, Germany 
\email{\{andrzej.dulny,andreas.hotho,anna.krause\}@informatik.uni-wuerzburg.de}}

\maketitle              
\appendix
\section{Toy Data Sets}
\subsection{Equation Systems}
The selected equations are listed in \Cref{tab:selected_equations} for easier overview.
The number of equations refers to the number of coupled equations in the PDE system.
For Burgers' equation and the gas dynamics equations the number of equations depends on the number of spatial dimensions as the velocity has one component per dimension.

\begin{table}[ht]
    \centering
    \caption{Summary of the selected equations showcasing their variety, including linear and non-linear equations, first and second-order equations and number of coupled equations.}
    \label{tab:selected_equations}
    \begin{tabular}{lcccccc}
    \toprule
    Equation & Linear  & Order & No. of eqns. \\ \midrule
    Advection-Diffusion (AD) & Yes & 1 & 1 \\
    Wave (W) & Yes & 2 & 1 \\
    Burgers' (B) & No & 1 & 2 \\
    Gas Dynamics (GD) & No & 1 & 4 \\
    \bottomrule
    \end{tabular}
\end{table}

\paragraph{Advection-Diffusion Equation.}
The advection equation
\begin{equation}
    \frac{\partial u}{\partial t} = - \nabla\cdot (\mathbf{c}u) +D\nabla^2 u
\end{equation}
describes the transport of a quantity described by a scalar field $u$ in a medium moving with the velocity $\mathbf{c}$ and the diffusion of a quantity from regions of higher concentration to regions of lower concentration driven by the gradient in concentration.
$D$ denotes the diffusion coefficient of a medium assumed to be constant in the whole domain.

\paragraph{Wave Equation.}
The wave equation
\begin{equation}
    \frac{\partial^2 u}{\partial t^2} = \omega^2 \nabla ^2 u
\end{equation}
describes the propagation of a wave in a given space where $u$ represents the amplitude and $\omega$ represents the speed of propagation.
It is a linear, second-order PDE.

\paragraph{Burgers' Equation.}
Burgers' equation 
\begin{equation}
    \frac{\partial \mathbf{u}}{\partial t} = D \nabla ^2 \mathbf{u} - \mathbf{u}\cdot\nabla \mathbf{u}
\end{equation}
is a non-linear second order PDE that commonly describes phenomena in fluid mechanics.
The equation describes the speed $u$ of a fluid in space and time with $D$ representing the fluid's viscosity.

\paragraph{Gas Dynamics.}
In gas dynamics, the system of coupled non-linear PDEs
\begin{equation}
    \begin{split}
        &\frac{\partial \rho}{\partial t} = -\mathbf{v}\cdot\nabla\rho - \rho\nabla\cdot\mathbf{v}\\
        &\frac{\partial T}{\partial t} = -\mathbf{v}\cdot\nabla T - \gamma T\nabla\cdot\mathbf{v} + \gamma\frac{Mk}{\rho}\nabla^2 T\\
        &\frac{\partial \mathbf{v}}{\partial t} = - \mathbf{v}\cdot \nabla\mathbf{v} - \frac{\nabla P}{\rho} + \frac{\mu}{\rho}\nabla(\nabla\mathbf{v})\\
    \end{split}    
\end{equation}
describes the evolution of temperature $T$, density $\rho$, pressure $P$ and velocity $\mathbf{v}$ in a gaseous medium.
The equations directly correspond to the conservation of mass, the conservation of energy, and Newton's second law~\cite{Anderson1995}.
The parameters specify the physical characteristics of the gas, $\gamma$ being the heat capacity ratio, $M$ the mass of a molecule of gas, and $\mu$ the coefficient of viscosity.

\paragraph{Parameters.}
For each equation we set the parameters to reasonable values as
summarized in~\Cref{tab:parameters_equations}. 
We additionally scale the magnitude of the derivative for each equation by a fixed parameter.

\begin{table}[ht]
    \label{tab:parameters_equations}
    \centering
    \caption{Parameters used for data generation}
    \setlength{\tabcolsep}{0.5em}
    \begin{tabular}{lll}
    \toprule
    Equation & Parameters & Scale \\ \midrule
    Advection-Diffusion (AD) & $c_x = c_y = 1$, $D=0.001$ & 0.1 \\
    Wave (W) & $\omega = 0.1$ & 0.1 \\
    Burgers' (B) & $D=0.01$ & 0.01 \\
    Gas Dynamics (GD) & $\gamma = 1$, $M = 1$, $\mu = 0.01$, $k = 0.01$ & 0.002 \\
    \bottomrule
    \end{tabular}
\end{table}

\subsection{Data generation}
We generate data by solving each of the selected PDE systems using a high-resolution discretization to obtain highly accurate dynamical data which is then sampled at a low resolution grid.
With this, we simulate real world dynamical data (e.g. weather data) being measured at a limited spatial resolution, while the underlying physical process is continuous.
 
Similar to~\cite{Karlbauer2019} the initial condition for each equation at $t=0$ is set as a sum of $N$ Gaussian bell curves
\begin{equation}
    \tag{1D case}
    u_0(x) = \sum_{i=1}^N a_i\exp(-(x-\mu_i)^2)
\end{equation}
\begin{equation}
    \tag{2D case}
    u_0(x,y) = \sum_{i=1}^N a_i\exp(-(x-\mu_i)^2-(y-\nu_i)^2)
\end{equation}
where $a_i \sim \mathcal{U}(-1,1)$ and $\mu_1, \nu_i \sim \mathcal{U}(-5,5)$. 

\begin{figure}[ht]
    \centering
    \includegraphics[width=\textwidth]{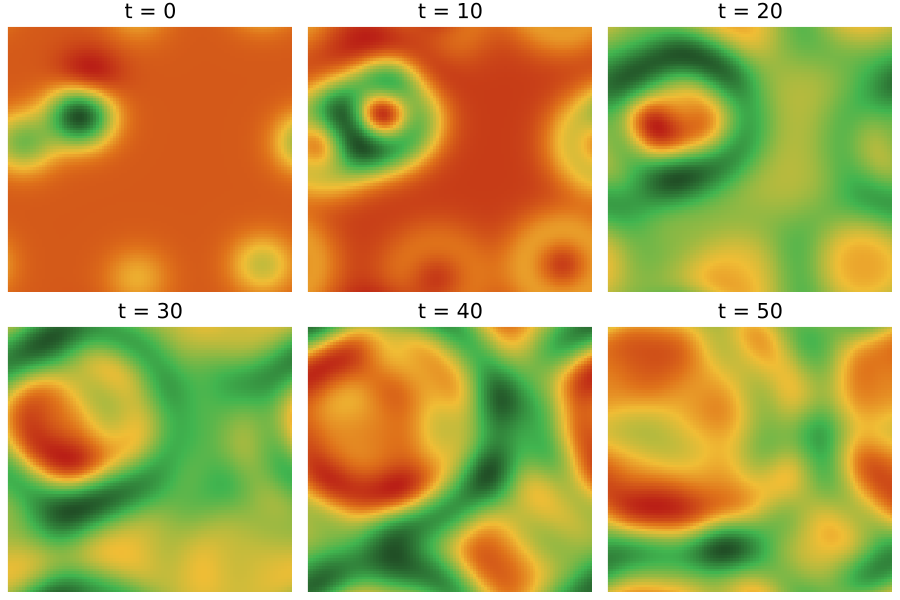}
    \caption{Example of a wave equation in two dimensions. Each picture shows the amplitude over the spatial domain at the given point in time.}
    \label{fig:wave_example}
\end{figure}

The spatial domain is set to $\Omega_2 = [0, 1]\times[0, 1]$ with $\Delta x = \Delta y = 0.01$ for the two dimensional case with periodic boundary conditions.

We solve the underlying initial value problem using a Method of Lines based PDE solver for $t\in [0, 501]$ with solutions saved at $\Delta t = 0.1$.
The ODE solver used for the Method of Lines is \textit{VCABM}, an adaptive order adaptive time Adams Moulton solver using an order adaptivity algorithm derived from Shampine's DDEABM~\cite{Rackauckas2020}. 
The generated high-resolution solutions are then sampled at a lower resolution of $\Delta x = \Delta y = 0.1$ and $\Delta t = 1$.

The training datasets are generated by solving the corresponding PDE system for \num{50} different initial conditions, initialized with $N=5$ independently sampled gaussian curves. 
Both the validation and test datasets are generated for \num{10} different initial conditions with $N = 5$.

\section{Further Plots}
\begin{figure*}[ht]
    \centering
    \includegraphics[width=0.9\textwidth]{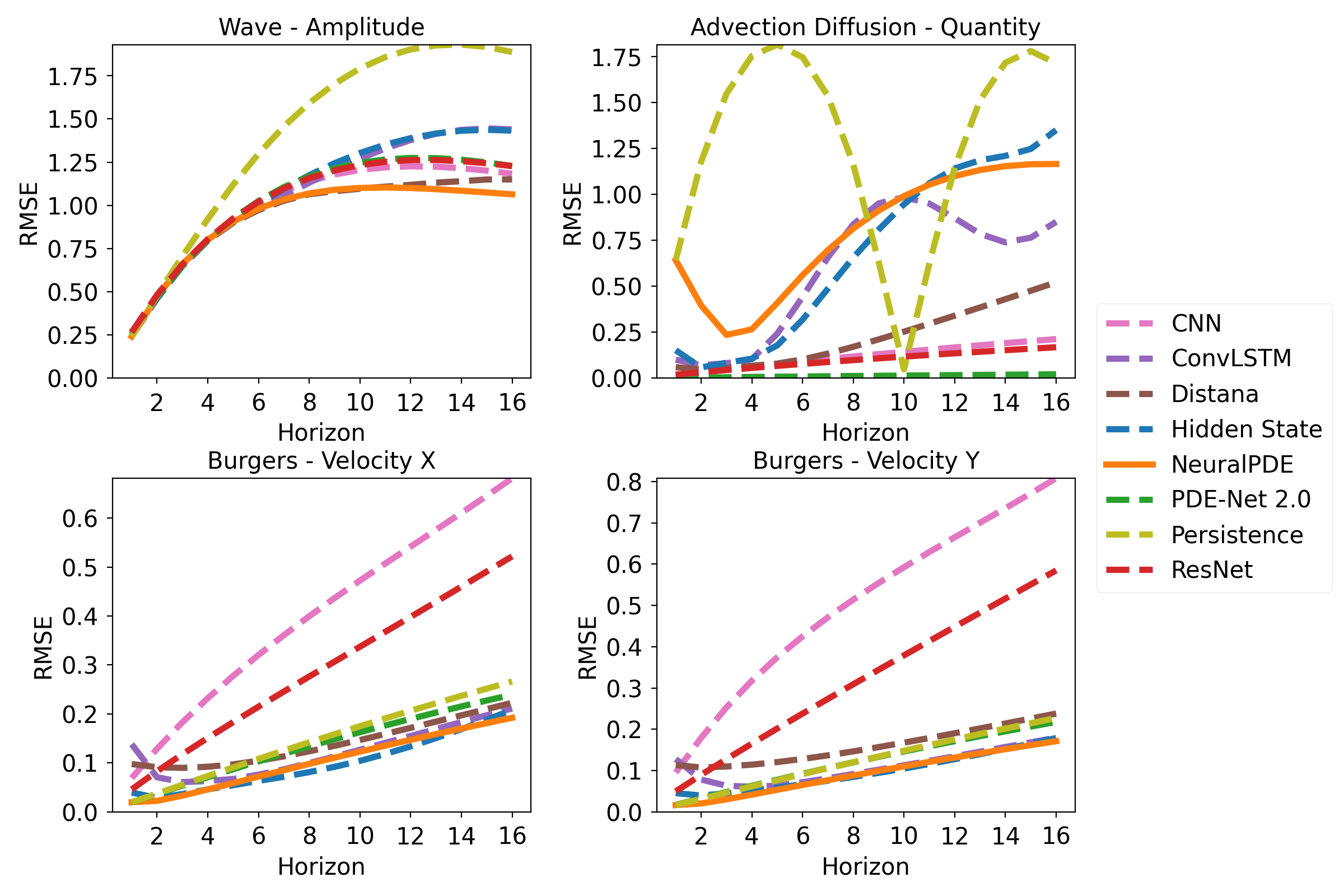}
    \caption{Predictions over different horizons. The figure shows the RMSE for the Burgers, Wave and Advection-Diffusion datasets for all tested models as a function of the prediction horizon.}
    \label{fig:predictions_horizon}
\end{figure*}
\begin{figure*}[ht]
    \centering
    \includegraphics[width=0.9\textwidth]{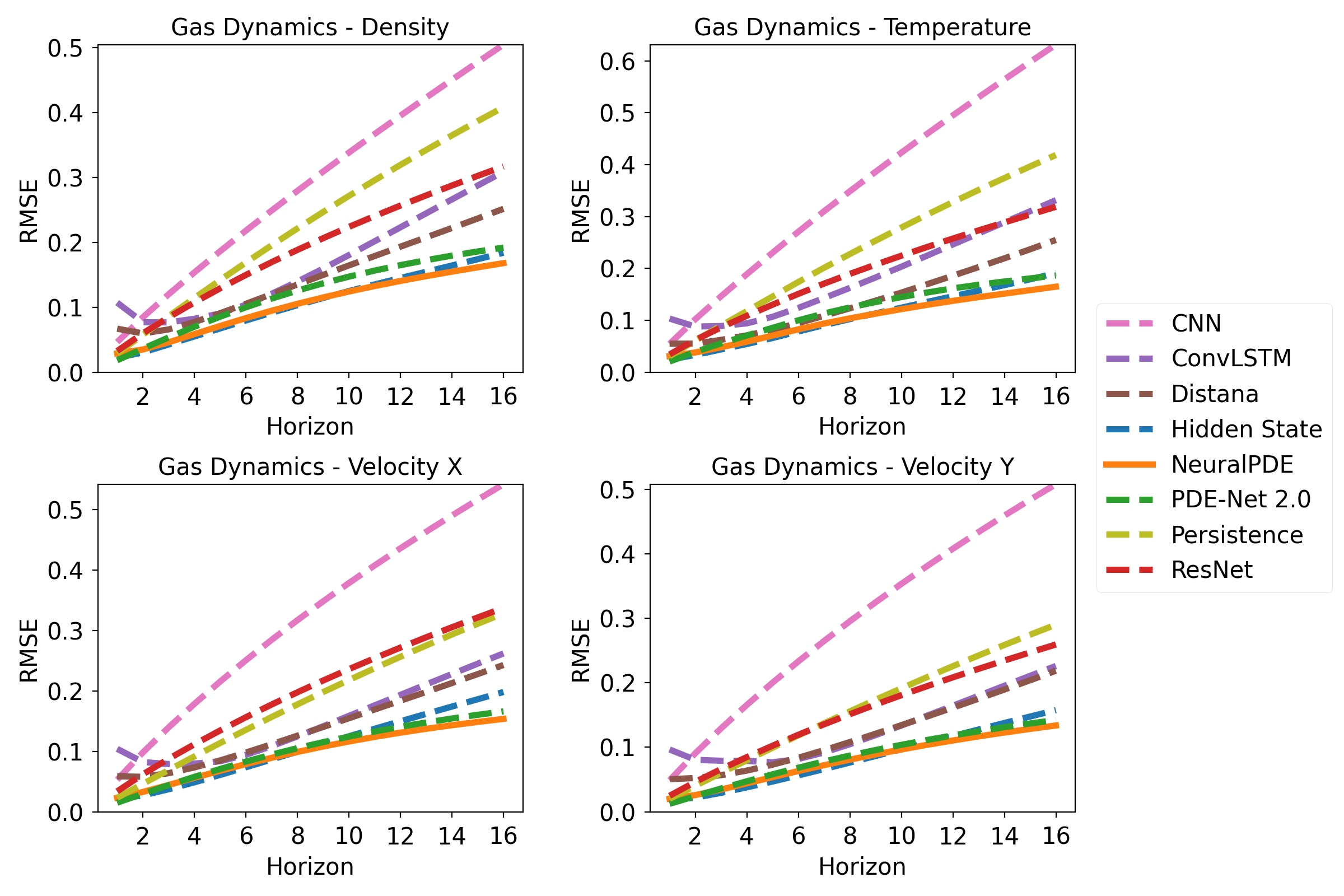}
    \caption{Predictions over different horizons. The figure shows the RMSE for the Gas Dynamics dataset for all tested models as a function of the prediction horizon.}
    \label{fig:predictions_horizon}
\end{figure*}
\begin{figure*}[ht]
    \centering
    \includegraphics[width=0.9\textwidth]{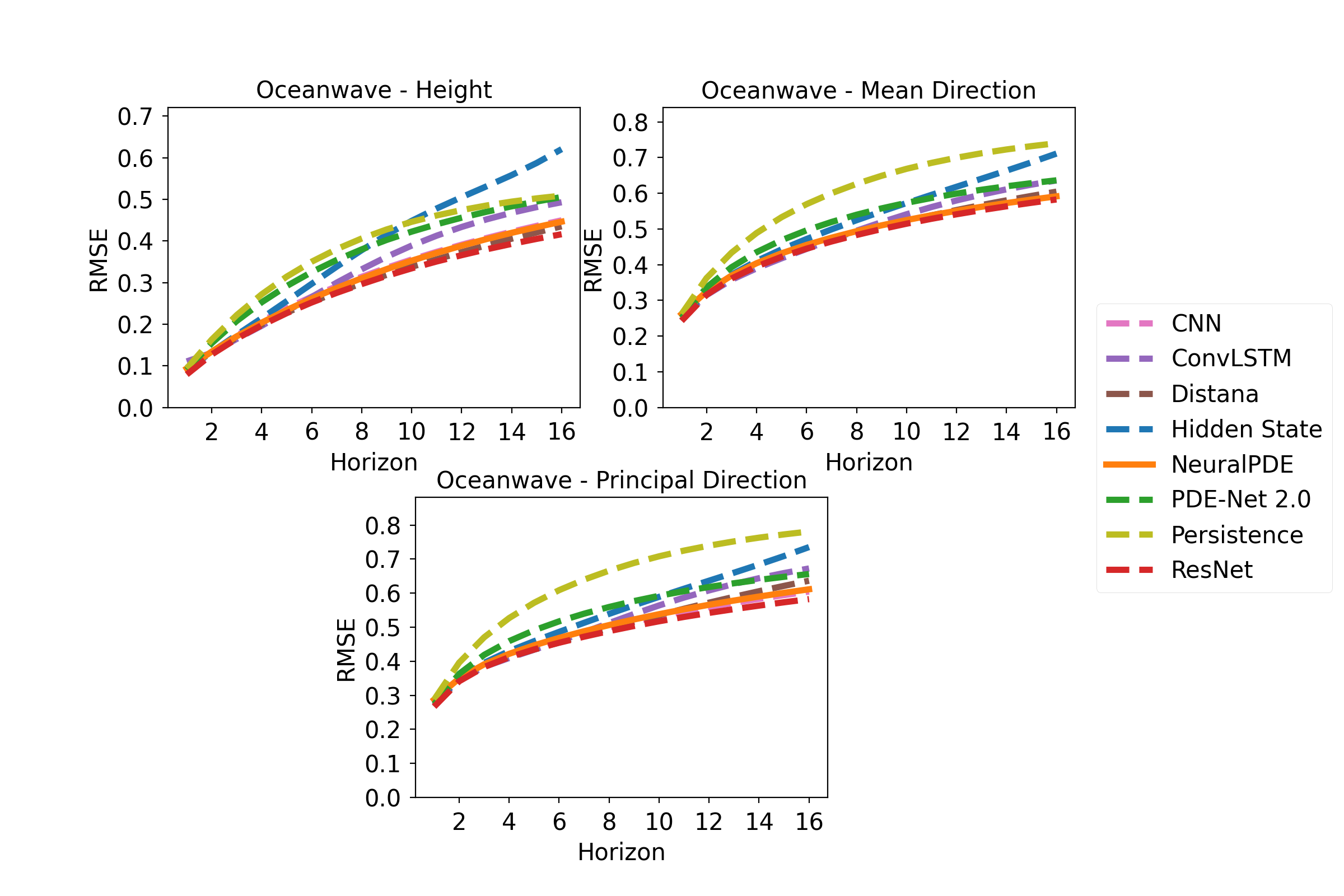}
    \caption{Predictions over different horizons. The figure shows the RMSE for the Ocean Wave dataset for all tested models as a function of the prediction horizon.}
    \label{fig:predictions_horizon}
\end{figure*}
\begin{figure*}[ht]
    \centering
    \includegraphics[width=0.9\textwidth]{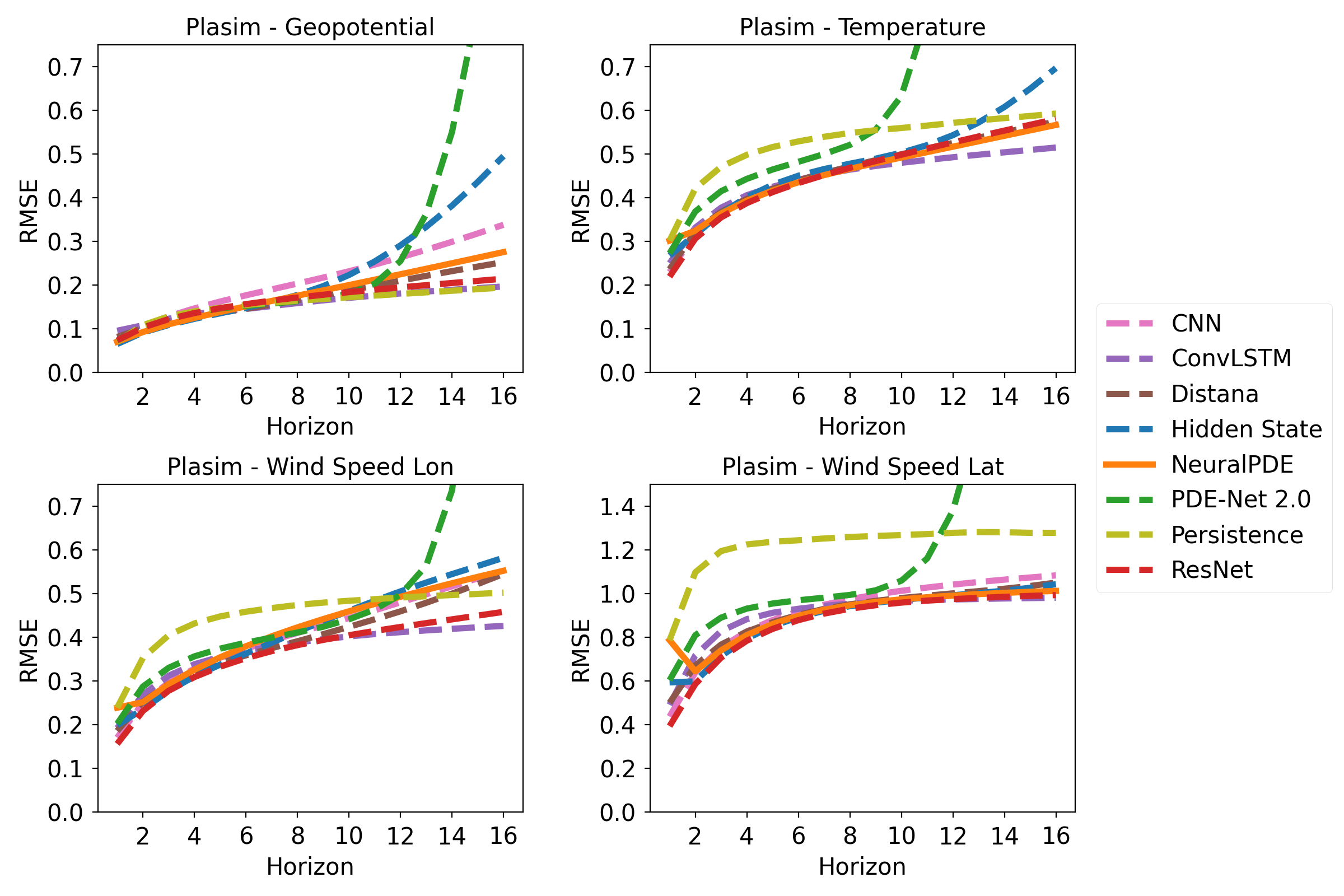}
    \caption{Predictions over different horizons. The figure shows the RMSE for the PlaSim dataset for all tested models as a function of the prediction horizon.}
    \label{fig:predictions_horizon}
\end{figure*}

\bibliographystyle{splncs04}
\bibliography{references}